\documentclass[10pt,twocolumn]{article}

\usepackage[T1]{fontenc}
\usepackage[utf8]{inputenc}
\usepackage{lmodern}
\usepackage{microtype}
\usepackage{amsmath,amssymb}
\usepackage{graphicx}
\usepackage{booktabs}
\usepackage{longtable}
\usepackage{subcaption}
\usepackage{dblfloatfix}
\usepackage{balance}
\usepackage{xurl}
\usepackage[hidelinks]{hyperref}
\hypersetup{
  pdftitle={LLMs Interpret, Embeddings Organize, Graphs Emerge: Agent-Driven Compilation of Scientific Knowledge},
  pdfauthor={Shi-Ju Ran, Kun Zhang, Xi Wu, Liu-Si Yang, Wen-Jun Li}
}
\usepackage[letterpaper,top=0.65in,bottom=0.75in,left=0.65in,right=0.65in,columnsep=0.25in]{geometry}
\newcommand{\safeincludegraphics}[2][]{%
  \includegraphics[#1]{#2}%
}
\newcommand{\cliopt}[1]{\texttt{\char45\kern0pt\char45#1}}
\newcommand{\doi}[1]{\href{https://doi.org/#1}{\nolinkurl{doi:#1}}}

\title{LLMs Interpret, Embeddings Organize, Graphs Emerge: Agent-Driven Compilation of Scientific Knowledge}
\author{%
Shi-Ju Ran\textsuperscript{1,*},
Kun Zhang\textsuperscript{1},
Xi Wu\textsuperscript{1},
Liu-Si Yang\textsuperscript{1}, and
Wen-Jun Li\textsuperscript{2}\\[0.45em]
\small\textsuperscript{1}Center for Quantum Physics and Intelligent Sciences, Department of Physics,\\[-0.1em]
\small Capital Normal University, Beijing 100048, China\\
\small\textsuperscript{2}College of Artificial Intelligence, Putian University, Putian, Fujian 351100, China\\[0.2em]
\small\textsuperscript{*}Corresponding author. Email: \href{mailto:sjran@cnu.edu.cn}{sjran@cnu.edu.cn}
}
\date{}

\begin{document}

\twocolumn[
\begin{@twocolumnfalse}
\maketitle
\begin{abstract}
Sustained scientific work requires a knowledge substrate that carries interpretation across tasks and preserves paths to source evidence. We call this process \emph{scientific knowledge compilation} and implement it in ASKS, the \emph{Agent-Driven Scientific Knowledge System}. For each source, an LLM produces a readable Wiki view and machine-facing semantics. Deterministic checks convert the latter into a document-local GraphDelta, and embedding geometry together with explicit graph rules integrates the proposed changes into persistent state. Each ingest is an inspectable state transition over accumulated knowledge, with compiled Wiki and graph views linked to the preserved source record. We examine this process by chronologically compiling 56 published papers from one research program. Branch survival, cross-paper support, lineage, coverage, and churn yield a source-traceable author research portrait centered on tensor-network methods, with branches into quantum many-body research, tensor-network machine learning, and quantum-AI-oriented directions. In this run, higher-level Hub organization remains stable and low-churn. Canonical-node growth is predominantly additive. Graph-level measurements and navigation paths retain links to the source records from which they were compiled.
\end{abstract}
\vspace{0.7em}
\end{@twocolumnfalse}
]

\section{Introduction}
Scientific work accumulates more than documents. A researcher gradually builds a working organization of methods, claims, terminology, open questions, and connections across time. The challenge of connecting facts scattered across separate literatures was crystallized in Don Swanson's 1986 account of \emph{undiscovered public knowledge} \cite{swanson1986undiscovered}, and literature-based discovery (LBD) developed this observation into methods for connecting separated literatures \cite{henry2017lbd,smalheiser2017rediscovering}. Four decades later, scientific records are larger, more heterogeneous, and produced faster, while the organization that supports a sustained research program must span papers, terminology, methods, and time.

Scientific agents change what can be done about this old problem. They increasingly search literature, call tools, analyze data, generate hypotheses, and participate in experimental workflows \cite{wang2023scientific,boiko2023autonomous,bran2024chemcrow,lu2024aiscientist,gottweis2026coscientist,ghareeb2026robin}. Retrieval-augmented systems can assemble external evidence for one task \cite{karpukhin2020dense,lewis2020retrieval,asai2026openscholar}, typically as transient context. Never-ending language learning established an earlier precedent for continually extending a structured knowledge base that supports later learning \cite{carlson2010nell}. Repeated agent workflows introduce a second infrastructure objective: to \emph{accumulate, revise, and reorganize} prior interpretation so that later agents inherit a persistent knowledge substrate and build directly on earlier work. Retrieval determines what an agent can read for the present task. Compilation determines what later tasks can inherit.

Several research lines provide pieces of such a substrate. Knowledge graphs and scientific knowledge graphs make entities, relations, and scholarly claims machine-operable \cite{bernerslee2001semantic,hogan2021knowledge,zhong2023kgconstruction,hofer2024construction,ding2026scientifickg}. LLM--KG research maps complementary roles for graph-enhanced models, model-augmented graphs, and synergized systems, while construction systems automate extraction, canonicalization, schema induction, and graph enrichment \cite{pan2024roadmap,zhu2024llmkg,zhang2024edc,bai2026autoschemakg,lu2025karma}. Sentence-embedding methods provide scalable semantic spaces for similarity search and clustering \cite{reimers2019sbert,gao2021simcse}. Dynamic-network and dynamic-KG studies characterize community birth, split, merge, and other forms of structural evolution \cite{dakiche2019community,aparicio2024dynamic}. These components are individually well established. A remaining architectural question is how to combine them in a persistent, source-linked \emph{source-to-graph state transition} whose document-local changes, accumulated construction history, and derived navigation surfaces can be inspected, replayed, and measured.

We formulate that transition as \emph{scientific knowledge compilation} and instantiate it in ASKS, the \emph{Agent-Driven Scientific Knowledge System}. An LLM encodes each incoming source as a human-readable Wiki view and a machine-facing semantic representation. A validated GraphDelta records the intended document-local changes before persistent writes. Embeddings place new and existing knowledge in a shared semantic geometry, and explicit identity, membership, and lifecycle rules convert similarity into admissible graph updates. Repeating the process yields an evolving state trajectory
\[
D_1,\ldots,D_t \xrightarrow{\mathrm{ASKS}} G_1,\ldots,G_t.
\]
The graph records the collective state produced by repeated compilation. Here \emph{knowledge compilation} is used in a source-transformational sense. Classical knowledge compilation maps formal theories into tractable target languages, whereas ASKS compiles scientific records into source-linked representations and evolving graph state \cite{darwiche2002map}. We define \emph{scientific knowledge compilation} as the repeated, provenance-preserving transformation of scientific records into revisable representations and persistent relational state that subsequent research tasks can inherit.

LLM-based interpretation can move much of the labor-intensive semantic extraction and ontology population into an auditable compilation stage, a motivation shared by recent automated knowledge-graph construction and enrichment systems \cite{zhang2024edc,bai2026autoschemakg,lu2025karma}. In ASKS, model output remains source-linked and passes explicit validation before persistent writes. Semantic embedding provides a reusable organizing geometry, while source-linked evidence carries scientific authority. Here graph emergence is a \emph{construction-level property}: higher-level navigational organization develops through repeated source-local compilation and graph-state transitions rather than arriving as a complete global structure from any individual source. This usage follows the local-interaction view of self-organization, where macroscopic structure arises from repeated lower-level interactions under a compact set of local rules \cite{gershenson2025selforganization}. It also connects to emergent-semantics research, where global semantic organization develops through repeated local interactions and graph transformations \cite{aberer2004emergent,mika2007ontologies}.

\paragraph{Claims and scope.} We make two methodological contributions. First, we formulate persistent scientific knowledge construction as a provenance-preserving compilation process in which each source induces an inspectable state transition over accumulated knowledge. Second, ASKS implements this formulation through validated document-local GraphDeltas, provenance-preserving fusion, embedding-based organization, persistent graph state, and replay. We examine the resulting construction trajectory on 56 chronologically ordered papers by measuring node reuse, multi-source support, membership churn, Hub birth, persistence, lineage, and portrait coverage. The present study evaluates formation and persistence within this trajectory. Order reconstruction, cross-domain transfer, and comparison with alternative graph-construction methods are separate robustness and comparative evaluations. Supplementary Sections~\ref{sec:supp-repro}--\ref{sec:supp-engineering} provide the frozen reproducibility record, complete organization and transaction rules, re-ingest and replay contracts, and implemented agent mechanisms. Supplementary Table~\ref{tab:supp-manifest} lists the complete 56-work corpus.

\begin{figure*}[!tbp]
    \centering
    \safeincludegraphics[width=0.98\textwidth]{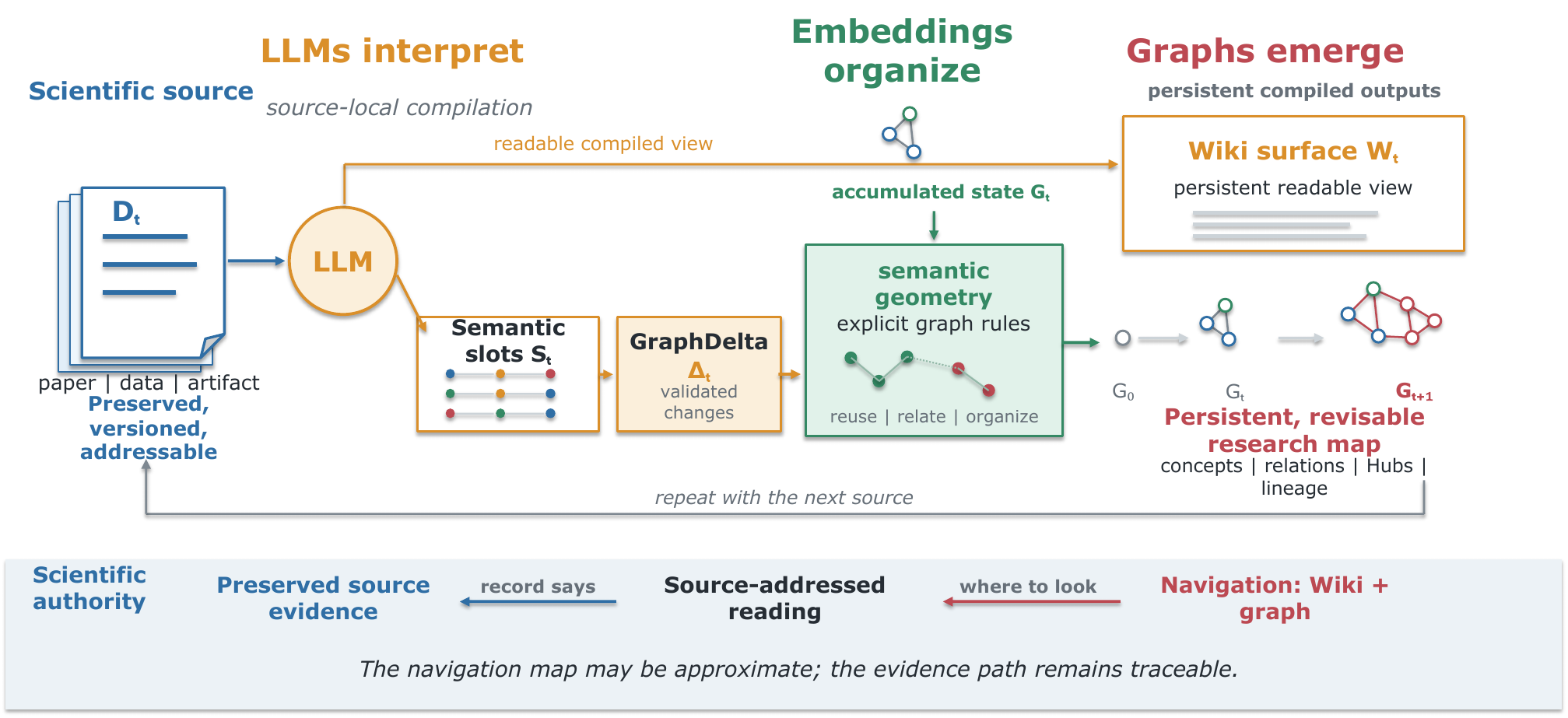}
    \caption{\textbf{Scientific knowledge compilation in ASKS.} Each received scientific source $D_t$ remains preserved while an LLM produces two complementary source-local representations: a readable Wiki view $W_t$ and machine-facing semantic slots $S_t$. Validated semantics together with deterministic source relations form an inspectable document-local GraphDelta $\Delta_t$. Embedding geometry and explicit graph rules integrate the delta with accumulated state $G_t$ to produce $G_{t+1}$. Repeating this transition turns local interpretation into an evolving research map. The persistent Wiki and graph surfaces guide navigation, while scientific claims resolve to source-addressed evidence.}
    \label{fig:conceptual-workflow}
\end{figure*}

Figure~\ref{fig:conceptual-workflow} summarizes how one preserved source becomes inherited scientific knowledge and how navigation remains connected to source evidence.

\section{ASKS: System Model}
\paragraph{From a scientist's perspective.} A researcher supplies papers or other scientific records to ASKS. Each received source remains preserved. ASKS compiles it into a readable view and machine-facing semantics, links the new material to accumulated knowledge, and updates an evolving research map. The map helps a researcher or agent decide where to look, while important factual paths lead back to the received sources. In practical terms, the output is an inspectable body of compiled notes and relations that can be revised and inherited across later research tasks.

ASKS separates what is received, what is compiled for use, and what is temporarily needed during construction. The \emph{source record} contains original papers or other research artifacts, stable metadata, and machine-readable derivatives used for addressing. The \emph{compiled knowledge} contains the persistent views that researchers and agents reuse: Wiki pages, keywords, propositions, graph relations, and Hubs. A \emph{Hub} is a persistent navigational region representing a coherent part of the accumulated research space. Its role is to organize where researchers and agents look, while factual authority remains with source evidence. A separate \emph{research state} contains open questions and provisional interpretations under review for possible later construction. These classes have different update rules because they play different scientific roles.

Between the source record and the persistent graph, an ingest also creates a transient \emph{compilation state}. This state includes semantic slots extracted from the current source, a validated document-local subgraph, and an in-memory graph delta that records what the ingest intends to attach or change. These objects are analogous to intermediate representations in a compiler: they make the transformation inspectable and remain transient construction intermediates. The persistent Wiki page and persistent graph are therefore sibling compiled products, while the semantic slots and graph delta guide their construction.

The core dataflow is
\[
D_t \xrightarrow{\mathrm{LLM}} \{W_t,S_t\} \xrightarrow{\mathrm{delta}} \Delta_t \xrightarrow{\mathrm{fusion+organization}} G_{t+1},
\]
where $W_t$ is the readable Wiki view, $S_t$ the machine-facing semantic representation, and $\Delta_t$ the validated document-local GraphDelta. Repeating this process yields the graph trajectory $G_0,G_1,\ldots,G_T$. The LLM interprets the current source. Global organization arises as successive local deltas interact with the accumulated graph.

\subsection{Source Record and the Two Compiled Surfaces}
Source handling follows a conservative rule: compilation creates convenient representations while preserving the received record. FAIR principles motivate persistent identification and reuse \cite{wilkinson2016fair}. PROV-DM separates entities, activities, agents, and derivations \cite{moreau2013provdm}. RO-Crate shows how heterogeneous research artifacts can be packaged with machine-readable metadata and relations \cite{soilandreyes2022rocrate}. In this system, these ideas become an engineering boundary. The source package is versioned and provides a stable reference while derived views are recreated or revised.

The first compiled surface is the \emph{Wiki view}. It is designed for people and language models to read: a short navigation summary and structured content linked back to the source. The second is the \emph{graph view}. It is designed for computation: reusable nodes and relations connect the Wiki page, raw package, concepts, people, propositions, and hubs. The graph view is stored separately from the Wiki page. Semantic slots are emitted through their own contract and compiled into graph state, so readable synthesis and machine-facing relational structure can evolve under different validation contracts.

Coupling human-readable documents with machine-processable semantics has a longer lineage. Semantic-document architectures connect documents and ontologies through multiple access forms, while Semantic MediaWiki places explicit semantic annotations in Wiki pages \cite{eriksson2007semanticdocument,krotzsch2007semanticwikipedia}. ASKS uses a different division of representation: the Wiki view and graph are separately validated sibling products compiled from the same preserved source. This common evidence root lets readable synthesis and machine-facing relations evolve under distinct contracts. The resulting plural representation is consistent with nanopublications, ORKG, and semantic-unit approaches that make scholarly knowledge addressable at different granularities \cite{kuhn2016nanopub,jaradeh2019orkg,vogt2024semanticunits}.

\subsection{The Three Computational Roles}
The slogan in the title summarizes one compiler across three scales. \emph{Knowledge compilation} refers to the complete source-to-graph transformation, with the LLM serving as its encoding stage. Table~\ref{tab:division-of-labor} summarizes the roles, outputs, and authority boundaries of the cooperating layers.

\paragraph{LLMs interpret.} The language model converts a human-oriented source into local semantics. It writes readable Wiki content and emits semantic slots describing source-local relations. This stage concentrates flexible language understanding in source-local output that passes validation before persistent graph writes. This source-local automation follows the broader move toward LLM-assisted knowledge-graph construction and enrichment, where manual extraction and schema population are increasingly delegated to models under explicit reconciliation steps \cite{zhang2024edc,bai2026autoschemakg,lu2025karma}.

\paragraph{Embeddings organize.} Compiled node profiles and Hub scopes are mapped into a shared semantic space, following the general use of sentence embeddings as reusable representations for semantic similarity and clustering \cite{reimers2019sbert,gao2021simcse}. Similarity in this space is reused throughout integration: identity disambiguation, proposition alignment, paper routing, node-to-Hub membership, and Hub lifecycle candidates. Embedding geometry is more than a retrieval index. It is the principal continuous variable coupling new local knowledge to the existing graph. Coded lexical gates, graph affinity, thresholds, hysteresis, routing checks, Hub capacity rules, and lineage constraints transform that continuous geometry into admissible discrete state changes.

\paragraph{Graphs emerge.} The graph records the accumulated consequence of repeated compilation. Direct source links and mechanical relations are explicitly compiled, while higher-level organization develops through the history of many ingests. Hub membership can relax as new nodes arrive. Overloaded Hubs can redistribute or refine. Regions awaiting Hub assignment can form new Hubs. Existing branches can split, merge, drift, or retire under lifecycle rules. The global configuration develops across the accumulated trajectory. These dynamics produce structural candidates, while an agent or researcher supplies the human-readable interpretation required for high-consequence events.

\subsection{Ingest as the State-Transition Boundary}
Throughout the paper, \emph{ingest} means the controlled state transition by which a source enters this system. It spans preprocessing, LLM encoding, semantic validation, local graph-delta construction, attachment planning, graph organization, validation, and logging. Fusion is transactional: either the complete document update succeeds or the previous graph state is restored. Tool-using LLM work provides the general pattern of combining model reasoning with external operations \cite{yao2023react,schick2023toolformer}. Here those operations have an additional role: they enforce an explicit contract before model output enters persistent scientific state.

This boundary separates learned judgment from mechanically checkable obligations. LLM generation may vary across runs, and embedding scores are model-derived numerical representations. Source identity, graph object types, hard delta errors, source links, write atomicity, and many postconditions can be checked programmatically. Operationally, source addressing, validation, duplicate removal, savepoints, origin recording, and snapshots are mechanical. Wiki writing, semantic slots, and embedding similarity are model-derived. Consequential identity decisions and Hub lifecycle commitments pass explicit gates. These decisions are recorded so that a later re-ingest or replay can identify which source and which operation changed the graph. Database provenance and truth-maintenance traditions provide useful precedents for retaining dependencies needed to explain and selectively revise derived state \cite{cheney2009provenance,doyle1979tms}.

\subsection{Authority and Interpretation}
The architecture separates computational organization from scientific authority. Sources provide primary evidence for what the received record states, while broader scientific assessment combines multiple records, analysis, and expert review. Wiki pages are readable compiled views. Semantic slots and graph deltas are construction intermediates. Graph nodes and Hubs provide persistent computational organization. Embedding scores quantify semantic proximity, while scientific correctness remains grounded in source evidence and judgment. Graph structure is used for navigation, whereas factual claims remain traceable to source evidence. Human judgment is concentrated where scientific meaning is assigned: resolving consequential ambiguity, approving a Hub scope, interpreting a split or merge, or deciding that an emergent organization merits investigation.

\begin{table*}[t]
\centering
\small
\setlength{\tabcolsep}{4pt}
\begin{tabular}{p{0.13\textwidth}p{0.32\textwidth}p{0.24\textwidth}p{0.22\textwidth}}
\toprule
\textbf{Layer} & \textbf{Principal role} & \textbf{Produced representation or signal} & \textbf{Scientific authority} \\
\midrule
Source record & Preserve the received scientific record and stable addressing. & Original files, metadata, and machine-readable derivatives. & Primary evidence for what the received record states. \\
LLM & Interpret one source and express local meaning. & Readable Wiki content and validated semantic slots. & Source-local interpretation under validation and review. \\
Embeddings & Place local and accumulated knowledge in a shared semantic geometry. & Similarity signals used by identity, routing, membership, and lifecycle gates. & Semantic proximity evaluated within coded gates. \\
Graph & Retain source-linked relations, Hub organization, and construction history. & A persistent, revisable navigation structure. & Source-linked navigation whose evidence resolves to source records. \\
\bottomrule
\end{tabular}
\caption{Division of labor and authority in ASKS. The four layers cooperate in one compiler and carry distinct scientific responsibilities.}
\label{tab:division-of-labor}
\end{table*}

\section{Knowledge Compilation: From Local Interpretation to Global Organization}
Knowledge compilation is implemented as a repeated \emph{ingest} protocol with two phases: \emph{encoding} and \emph{consolidation}. Encoding asks what one source says and produces source-local representations. Consolidation asks where those representations belong relative to accumulated knowledge. The first problem is primarily semantic and language-facing. The second is geometric, structural, and stateful. Their composition connects source-local interpretation to persistent graph organization.

\subsection{Encoding: One Source, Two Complementary Outputs}
Each source produces two complementary outputs under separate contracts. The first is a persistent Wiki page for human and model reading. A deterministic skeleton supplies metadata and section structure, while the LLM fills a short \emph{Navigation} summary and structured \emph{Content}. The pipeline supplies source paths deterministically. The Wiki therefore serves as a readable compiled view and traceback surface, with the received source retained as the factual reference.

The second output is a machine-facing semantic representation. The LLM emits semantic slots in a subject--predicate--object form. These slots are stored outside the Wiki and validated separately before graph construction. Readable synthesis can optimize for context and communication, while semantic slots can be normalized, type-checked, filtered, repaired, or rejected. The Wiki remains the readable interface and the semantic slots provide the graph-facing contract.

The two outputs pass different validation stages. Wiki validation checks required metadata and section structure. Semantic validation checks predicate registration, duplicates, citation fragments, bare abbreviations, and descriptive phrases. Hard violations block the transaction. Soft violations generate explicit warnings for later resolution or repair. The LLM therefore acts as a flexible interpreter under explicit persistent-write constraints.

\subsection{GraphDelta: The Document-Local Intermediate Representation}
Graph construction begins by combining three classes of relations. First, deterministic relations are derived mechanically from source metadata and addressing: the Wiki page is connected to its raw package, author metadata can create authorship edges, and declared related records can create citation links. Second, validated semantic slots contribute source-local semantic edges. Third, the compiler may create a small number of mechanical containment relations from already accepted nodes.

These relations first enter an in-memory \emph{GraphDelta} containing the page, raw packages, candidate edges, boundary mentions requiring identity resolution, canonical endpoints, and hard structural errors. In practical terms, GraphDelta is an inspectable list of graph changes proposed by the current source before those changes become persistent. It is the compiler's document-local intermediate representation.

The delta is checked before attachment. Every source requires a resolvable raw package. Empty triples and self-loops are hard errors. Exact duplicate triples are mechanically removed. Boundary mentions then pass through an ordered identity plan: canonical IDs and unique aliases take priority, followed by decomposed-name matches and finally an embedding-assisted identity gate. This gate decides whether a concept in the new source refers to an existing graph object or should remain distinct. It requires lexical identity evidence together with label and semantic embedding tests and a sufficient winner margin. An ambiguous candidate produces an explicit abstention and leaves the affected edge pending. Persistent reuse therefore requires sufficient identity evidence.

Topical routing illustrates the local/global boundary. Source-local semantic-slot predicates such as ``research keyword'' or ``main research'' remain local descriptors. Global paper routing compares the paper's research-positioning text with active Hub scopes in embedding space. The LLM interprets what the paper says, while the research map accumulates through cross-source Hub routing.

\subsection{Transactional Fusion and Provenance-Preserving Reuse}
After attachment planning, the document delta is fused inside a database savepoint. The complete document update either commits or returns to the previous graph state. Fusion creates or reuses nodes according to the plan, writes each distinct edge once, and then verifies that the page node, raw package, and source edge remain present. A hard-check violation triggers complete rollback to the savepoint and restores the prior complete state.

Cross-source reuse preserves lineage in a compact canonical topology. When an identical edge already exists, the new page is added to its origin record. Thus a relation can accumulate independent source contributions while remaining one reusable graph edge. This matters both scientifically and operationally: multi-source support appears as explicit lineage within stable topology, consistent with provenance-oriented representations of multi-source assertions \cite{menotti2025multisource}.

The compiler analogy maps directly onto the implementation. Semantic slots are a machine-facing intermediate form. GraphDelta is a planned object-level state change. Attachment behaves like linking against existing symbols. The savepoint is the commit boundary. The persistent graph is the accumulated output of validated compilation across sources and time.

This separation also clarifies the relation to recent LLM-based knowledge-graph construction. EDC makes canonicalization a distinct phase after extraction \cite{zhang2024edc}, while AutoSchemaKG and KARMA demonstrate increasingly autonomous schema induction and graph enrichment \cite{bai2026autoschemakg,lu2025karma}. Scientific-document systems make the source-local stage especially concrete: paper2lkg constructs a local graph from an individual academic paper, Oarga \emph{et al.} generate scientific ontologies and graphs from literature, and SciGraph-LLM transforms scientific PDFs into evidence-grounded, provenance-aware graphs under textual-fidelity constraints \cite{chen2025paper2lkg,oarga2026scientifickg,malashin2026scigraph}. ASKS takes source-grounded local construction as its starting point and extends it through incremental, stateful cross-document reconciliation: each document is linked against the graph produced by all previous ingests, proposed changes become inspectable before the write boundary, Wiki and graph persist as sibling compiled surfaces, and the resulting state trajectory is retained for replay and measurement.

\subsection{Embedding-Driven Organization}
Once the source-local delta has been fused, the graph is reorganized through a family of local rules that reuse the same embedding geometry. Numerical thresholds are specific to the demonstrated configuration. The frozen values are reported in Supplementary Table~\ref{tab:supp-thresholds}.

\paragraph{Identity and proposition alignment.} A nearest-neighbor score proposes candidates, while the configured multi-signal gate establishes node reuse. Claim-bearing propositions use a still more conservative policy: very high similarity reuses the proposition, an intermediate band records a semantic relation while preserving both nodes, and lower similarity leaves them independent. The middle band lets semantic proximity become explicit graph structure while preserving distinct claims.

\paragraph{Membership as continuous-to-discrete organization.} After each ingest, ordinary nodes in the affected neighborhood are reconsidered for Hub membership. For keywords and propositions, semantic affinity dominates and graph-neighborhood affinity contributes a structural signal. Membership uses hysteresis: a node needs stronger evidence to enter a Hub than to remain in it, preventing small score fluctuations from repeatedly reassigning the node. In addition, when parent and child Hubs are both plausible, the child receives a small preference, biasing the system toward the more specific organization. The memberships are rebuildable from current state and therefore remain revisable. Embedding geometry supplies a continuous interaction field. Thresholds, structure, and hysteresis turn that field into relatively stable discrete membership.

\paragraph{Hub birth, overload, split, and merge.} Nodes awaiting Hub assignment can form a birth candidate when their pairwise embedding graph contains a sufficiently large and cohesive component. Existing Hubs are also active dynamical objects. A Hub whose membership exceeds the configured capacity first redistributes members toward usable children. The remaining overloaded Hubs proceed to split analysis. Split acceptance requires separable member geometry, distinguishable child scopes, and routing probes that functionally recover their members. Merge candidates combine Hub-scope and prototype similarity with a recent-lineage eligibility constraint described next. In all three cases the numerical mechanism proposes the structure. An agent supplies or confirms the human-readable semantic commitment required for a high-consequence write.

\subsection{Lineage as Structural Memory}
Embedding similarity provides attraction, and lineage memory gives this attraction direction and persistence. After a split, explicit ancestry keeps newly refined branches distinct across subsequent ingests. The implementation therefore treats hierarchy as a state variable with memory.

A split creates explicit parent--child lineage. When a node is compatible with both a parent and a child Hub, the child receives a small membership preference, encouraging knowledge to settle into more specific regions. If a Hub grows beyond twenty active members, the dynamics first attempts to redistribute members into existing children. Remaining overloaded Hubs become new split candidates. This creates a directional pressure from broad overloaded regions toward finer organization.

The complementary rule is lineage retention. Merge analysis traces parent--child ancestry up to three generations. Two canonical Hubs whose recent ancestor sets intersect are treated as \emph{kin} and retain separate branch status. Branches with distinct lineages remain eligible for a lineage-preserving merge: the retired Hub is preserved through a redirect and its history remains addressable.

These rules give the organization dynamics two forms of memory. Hysteresis is \emph{temporal memory} at the membership boundary. Lineage is \emph{structural memory} at the hierarchy boundary. Together they make graph emergence a continuous dynamical process whose state carries forward across ingests.

\subsection{Compilation as a Replayable and Revisable State Transition}
A complete ingest is one ordered compiler pass: verify and preprocess the source, encode local meaning, validate the semantic outputs, build and inspect a document-local delta, fuse that delta transactionally, reorganize the affected graph neighborhood, and validate the resulting state. The ordering preserves causality: one source is interpreted before its local representation is attached globally, and the planned delta is visible before it becomes persistent state. The complete state machine and write invariants are given in Supplementary Section~\ref{sec:supp-state-machine}.

A useful knowledge compiler supports revision as well as growth. Re-ingest repeats construction for an existing source when its derivative, metadata, or local semantics must be reconsidered. Recompilation rebuilds derived state from recorded sources and dependencies. Hub membership is rebuildable, and lifecycle operations preserve lineage and history. These choices make the compiled surfaces revisable while maintaining a stable source boundary. Supplementary Sections~\ref{sec:supp-reingest} and~\ref{sec:supp-repro} describe the corresponding replay, isolation, and frozen-artifact contracts.

The compilation operator can be summarized as
\[
\Delta_t=C_{\mathrm{LLM}}(D_t), \qquad G_{t+1}=F(G_t,\Delta_t;E,\Theta),
\]
where $D_t$ is the newly ingested source, $C_{\mathrm{LLM}}$ denotes source-local semantic compilation into a validated document delta, $E$ is the embedding model, and $\Theta$ contains the identity gates, thresholds, graph-affinity rules, hysteresis, capacity limits, and lineage constraints. Section~5 studies the macroscopic behavior produced by repeatedly applying this compiler.

\section{Using the Compiled Structure: Navigation Before Answering}
Querying is secondary to compilation in this paper. It provides a practical test of whether the compiled state is useful to a scientist. Semantic and graph representations help the agent decide \emph{where to look}. Source-addressed reading establishes \emph{what the record says}. Information retrieval and RAG optimize access to useful context \cite{karpukhin2020dense,lewis2020retrieval,asai2026openscholar}, whereas scientific claim verification identifies support, contradiction, and questions requiring further evidence \cite{wadden2020scifact}. The system keeps those roles separate.

\subsection{Retrieval over the Same Semantic Geometry}
The embedding space used during compilation also provides a retrieval surface. A scientist can ask a normal-language question. Semantic retrieval proposes related keywords, propositions, pages, or source regions even when the wording differs. The graph contributes complementary structure through explicit relations, Hubs, and neighborhoods \cite{hogan2021knowledge,edge2024graphrag}. Similarity and graph position allocate attention, while source-addressed evidence provides factual authority.

Compiled pages then act as reading indexes and source locators. An agent can inspect a short lead and heading structure, choose the relevant section, and follow references to stable source locators. This staged reading concentrates relevant context and makes the scope of reading explicit. Long-context reliability varies with position in large inputs \cite{liu2024lostmiddle}. A precise source operation can therefore distinguish ``the document was available'' from ``this identified passage was actually read.''

\subsection{Support Traces and Provisional Research State}
The internal output of a query is a \emph{support trace}. A source-supported claim names the source region that bears on it. A composite judgment combines several source-supported claims and records its dependencies. An open item records questions requiring further retrieval together with conflicts and explored paths when available. This is related to citation-aware and post-hoc attribution work \cite{gao2023citations,gao2023rarr}. Here the trace derives directly from the operations that performed retrieval and reading, providing process-level provenance.

A composite judgment belongs to provisional research state, while Section~5 studies graph emergence in persistent compiled organization. An explicit construction or review operation can later promote a useful hypothesis into a permitted persistent representation. A substrate-level search result describes the coverage of this corpus. Broader scientific status draws on external review. For natural-science researchers, the ordinary interaction of asking a question and reading an explanation is backed by a record of what was retrieved, what was read, and which questions remain open.

\section{Graphs Emerge: From 56 Papers to an Author Research Portrait}
Graph emergence names the macroscopic behavior produced when the compiler in Section~3 is applied repeatedly. Each source creates a local delta $\Delta_t$. Fusion and organization transform the previous state $G_t$ into $G_{t+1}$. We analyze the trajectory
\[
G_0\rightarrow G_1\rightarrow\cdots\rightarrow G_T.
\]
The phrase \emph{graphs emerge} in the title refers to structure observed along this evolving trajectory after many local interpretations have been integrated.

Node reuse and creation occur at the attachment level, Hub membership at the mesoscopic level, and hierarchical refinement and lifecycle events at the organizational level. We use \emph{graph emergence} for persistent higher-level navigation that arises from repeated updates under local compilation rules, consistent with the local-to-global framing of self-organizing systems \cite{gershenson2025selforganization}. Its evidential scope is the navigational organization observed along the compiler's state trajectory.

Dynamic-network research already studies community birth, split, merge, growth, and death \cite{dakiche2019community}, and dynamic knowledge graphs have been used to identify emerging communities in scientific corpora \cite{aparicio2024dynamic}. ASKS uses a source-compilation mechanism: incoming graph elements are first interpreted from individual scientific sources by an LLM, then integrated through a persistent embedding geometry, provenance-preserving reuse, membership relaxation, capacity pressure, and hierarchy-aware lineage rules.

Agentic graph-expansion work provides a close complementary comparison. Buehler couples a reasoning-native LLM to a continually updated graph, recursively generates concepts and relations, and observes hubs, modularity, and bridges in a self-organizing knowledge network \cite{buehler2025agentic}. ASKS grows through a source-grounded compilation trajectory: preserved scientific records produce source-local candidate changes, validation gates persistent writes, and accepted deltas accumulate across documents. Emergent organization therefore serves as traceable navigation over scientific records, while scientific novelty is assigned through review.

\subsection{How Local Updates Become Global Structure}
At the attachment level, a source-local mention may reuse an existing node, remain ambiguous, or create a new node. Reused edges accumulate source origins within a compact canonical topology. At the Hub level, affected nodes are re-evaluated using semantic and structural affinity. At the hierarchical level, capacity and lineage rules allow coherent unassigned regions to form new Hubs and existing organization to refine while preserving its history. The final research map accumulates through these updates.

Embedding similarity is the principal continuous variable coupling new knowledge to existing state. For profiles $u$ and $v$, the basic semantic interaction is
\begin{equation}
  s(u,v)=\frac{\langle E(f(u)),E(f(v))\rangle}
  {\|E(f(u))\|\,\|E(f(v))\|},
\end{equation}
where $f$ is the textual profile and $E$ the embedding model. This equation supplies a shared geometry that is reused across several local rules, each adding its own lexical, structural, or lifecycle constraints.

The resulting dynamics have two explicit forms of memory. Membership hysteresis provides \emph{temporal memory}: stronger evidence is required to enter a Hub than to remain. Hub lineage provides \emph{structural memory}: ancestry guides later refinement and stabilizes newly split families as distinct branches. The complete lifecycle rules are reported in Supplementary Sections~\ref{sec:supp-thresholds}--\ref{sec:supp-state-machine}.

\subsection{What We Mean by ``Graphs Emerge''}
We distinguish three evaluation levels. \emph{Formation} means that repeated source-local compilation produces a persistent higher-level navigation structure under local rules. \emph{Robustness} asks whether a functionally equivalent structure reappears under reconstruction or perturbation. \emph{Scientific interpretation} asks what the structure means to a domain researcher. A Hub can support navigation at the formation level, while robustness and interpretation provide subsequent evaluation layers.

The worked example below demonstrates formation, multi-source support, and persistence along one frozen chronological trajectory. Its evaluated scope is this trajectory. Order robustness, causal decomposition, and domain interpretation form distinct subsequent evaluation layers.

\subsection{Worked Longitudinal Demonstration}
\paragraph{Corpus and compilation boundary.} The demonstration uses one deliberately bounded and physically isolated longitudinal corpus. A frozen manifest begins with 65 candidate records from one research program and applies fixed inclusion rules to obtain 56 independently published works with verified canonical PDFs, three excluded candidates, and six related-only records. Formal journal issue year is used for chronological grouping when it differs from an online-first date. The included works span 2010--2026 and are grouped by the frozen formal-publication year. The complete bibliographic manifest is provided in Supplementary Table~\ref{tab:supp-manifest}.

This corpus offers an interpretable demonstration boundary centered on one research program: the source set is explicit, chronology is meaningful, and the resulting organization can be compared with a known research trajectory. The production knowledge base sits outside the run. Local compilation, fusion, Hub pages, embedding caches, and graph snapshots occupy a physically isolated workspace.

\paragraph{Frozen two-phase construction.} Starting from clean derived state, the 56-paper corpus is compiled in the frozen chronological order defined by the publication manifest. Source-local LLM compilation is completed and frozen before graph-dependent fusion, so the trajectory shows how a fixed set of local interpretations is attached and organized as accumulated state grows. After every source, the run records identity decisions, multi-source lineage, Hub memberships, Hub lifecycle events, and a complete graph snapshot.

Phase L freshly extracts and locally compiles every included paper into a Wiki view, semantic slots, and a validated source-local bundle. After structural and stratified semantic audits, these 56 bundles are frozen. Phase G then begins from an empty derived graph $G_0$ and fuses the bundles in the frozen manifest order, grouped by formal publication year. Identity attachment, proposition alignment, membership refresh, and Hub lifecycle analysis are evaluated against the current $G_{t-1}$, while local LLM output remains fixed. Every candidate Hub birth requires a complete hash-bound commit or reject decision. The frozen models, thresholds, code hashes, and isolation paths are reported in Supplementary Section~\ref{sec:supp-repro}.

The run completed all 56 fusion steps and generated all 57 database snapshots. Every snapshot passed a read-only integrity check. All 170 artifact receipts matched their recorded SHA-256 values. All 56 step reports passed graph validation. The construction workspace remained isolated from the production Wiki and graph.

To instrument the compilation trajectory, we record three observables. For the eligible attachment decisions of paper $t$, the reuse fraction is
\begin{equation}
R(t)=\frac{n_{\mathrm{reuse}}(t)}
{n_{\mathrm{reuse}}(t)+n_{\mathrm{create}}(t)}.
\end{equation}
For the trajectory through step $t$, cumulative reuse is
\begin{equation}
R_{\mathrm{cum}}(t)=
\frac{\sum_{\tau=1}^{t} n_{\mathrm{reuse}}(\tau)}
{\sum_{\tau=1}^{t}[n_{\mathrm{reuse}}(\tau)+n_{\mathrm{create}}(\tau)]}.
\end{equation}
Abstentions are reported separately, and both denominators contain reuse and create decisions. Let $\mathrm{Supp}_t(v)$ be the set of distinct work identifiers supporting eligible canonical node $v$ after step $t$, let $N_{\geq2}(t)=\#\{v:|\mathrm{Supp}_t(v)|\geq2\}$, and let $N_{\mathrm{eligible}}(t)$ be the number of eligible canonical knowledge nodes. Multi-source consolidation is
\begin{equation}
M(t)=\frac{N_{\geq2}(t)}{N_{\mathrm{eligible}}(t)}.
\end{equation}
Finally, let $V_{t-1}$ be the eligible ordinary nodes already present before paper $t$, and let $H_t(v)$ denote the active Hub-membership set of $v$. Old-node membership churn is
\begin{equation}
C(t)=\frac{\#\{v\in V_{t-1}:H_t(v)\neq H_{t-1}(v)\}}{|V_{t-1}|}.
\end{equation}
Thus $C(t)$ measures reassignment of previously present nodes, while a new node's first membership is treated as growth.

\begin{figure*}[!tbp]
    \centering
    \safeincludegraphics[width=0.82\textwidth]{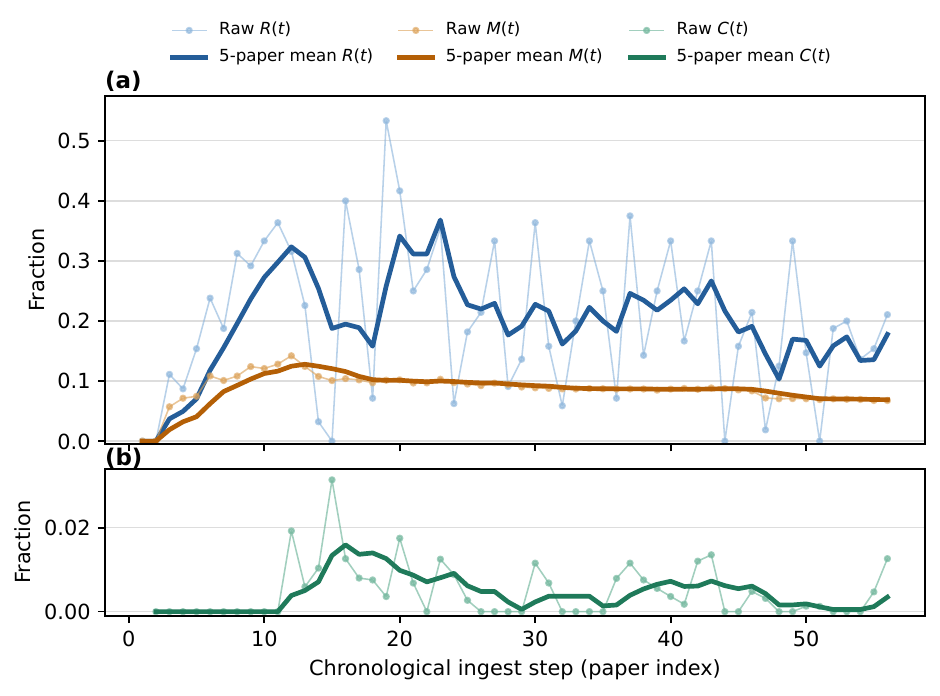}
    \caption{\textbf{Chronological compilation trajectory in the 56-paper demonstration.} (a) Canonical reuse $R(t)$ and multi-source consolidation $M(t)$ across chronological ingests. (b) Old-node Hub-membership churn $C(t)$ shown on its own scale because its amplitude is much smaller. Light marker-lines are raw per-paper values, whereas dark curves are five-paper trailing means used as visual guides. The diagnostics show low churn throughout and slightly declining descriptive trends in reuse and consolidation.}
    \label{fig:trajectory}
\end{figure*}

Across the run, 204 eligible attachment decisions reuse a canonical node, 878 create a node, and 5 abstain, giving $R_{\mathrm{cum}}(56)=0.1885$. The mean per-paper $R(t)$ is 0.2043, while the first-ten and last-ten means are 0.1715 and 0.1512. The descriptive ordinary-least-squares slope is $-5.41\times10^{-4}$ per paper, indicating a slightly declining reuse trajectory.

The multi-source consolidation fraction $M(t)$ peaks at 0.1420 at $G_{12}$ and finishes at 0.06764. Cross-source support accumulates in absolute terms: the numerator grows from 24 multi-source nodes at the peak to 60 at $G_{56}$, while the eligible-node denominator grows faster, from 169 to 887. Its descriptive slope is $-2.94\times10^{-4}$ per paper. In this compiler, node creation outpaces consolidation and produces a declining consolidation fraction within the expanding canonical vocabulary.

\begin{figure*}[!tbp]
    \centering
    \safeincludegraphics[width=0.94\textwidth]{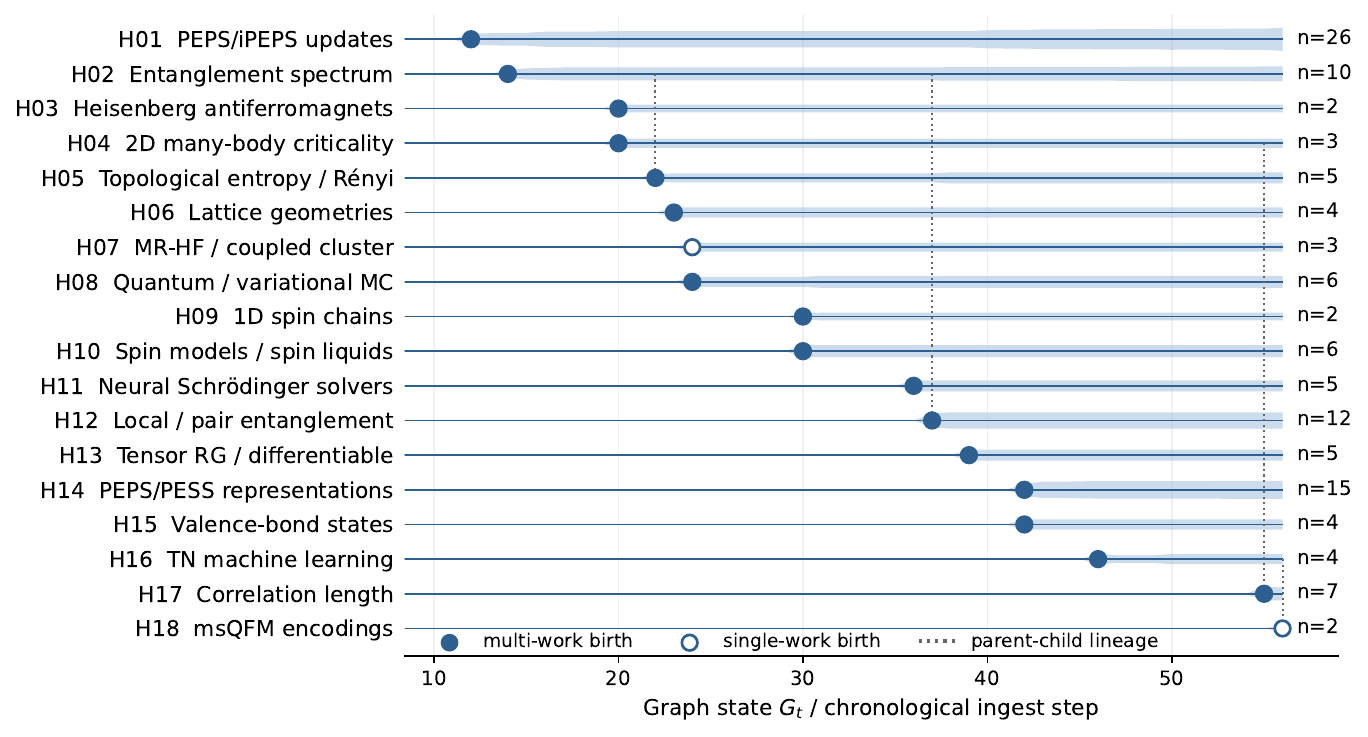}
    \caption{\textbf{Hub births, growth, and lineage during chronological compilation.} Ribbon width encodes active Hub membership over graph state $G_t$. Filled and open circles mark multi-work and single-work births, dotted links mark committed parent--child lineage, and right-side labels report final member count at $G_{56}$. The lifecycle consists of 18 births. Split, merge, and retirement counts are each zero.}
    \label{fig:lineage}
\end{figure*}

At the mesoscopic scale, old structure changes little. Mean $C(t)$ is 0.00467, the median is 0.00131, 26 of 55 defined steps have zero old-node membership change, and the maximum is 0.03139. The lifecycle is purely birth-driven. The chronological run commits 18 Hub births, and all 18 remain active at $G_{56}$. The split, merge, and retirement counts are each zero. ASKS implements all four lifecycle operations. This trajectory empirically exercises Hub birth and membership dynamics, while split, merge, and retirement remain untriggered. Of the 18 birth events, 16 are already multi-work at birth and four create parent--child lineage. Here a multi-work birth means that the committed birth-event record contains at least two distinct contributing work IDs. This event-level classification differs from the supporting-paper count obtained by taking the union of source papers attached to the Hub's active member nodes for visualization. The two single-work births remain valid navigation units that capture focused source-local structure. Figures~\ref{fig:trajectory} and~\ref{fig:lineage} therefore show a birth-dominated, low-churn navigation layer with cross-work structure present from the outset.

The 2024 density-outlier paper, D047 in the frozen sequence, is ``Eigenstate thermalization and its breakdown in quantum spin chains with inhomogeneous interactions'' (DOI: \href{https://doi.org/10.1103/PhysRevB.109.045139}{10.1103/PhysRevB.109.045139}). It contributes 105 create decisions corresponding to 104 unique newly created canonical nodes, 2 reuse decisions, and 11.96\% of all create decisions. A deterministic direct-contribution sensitivity analysis on the frozen formal outcomes raises $R_{\mathrm{cum}}(56)$ from 0.18854 to 0.20718 and final $M$ from 0.06764 to 0.07663 when D047's direct contribution is removed. The corresponding $R(t)$ slope is $-3.09\times10^{-4}$ per paper. This direct-contribution analysis preserves the downstream fusion sequence and all other frozen outcomes. These results position the density outlier as an amplifier within a broader additive-vocabulary trajectory. Supplementary Figure~\ref{fig:supp-sensitivity} reports the complete sensitivity traces.

The worked demonstration shows \emph{structural navigation compilation}: fixed local interpretations accumulate into a persistent, low-churn Hub organization along one real chronological sequence. Canonical consolidation follows a slightly declining trajectory, while order robustness defines a subsequent reconstruction study. In this frozen trajectory, \emph{the graph organizes before the vocabulary consolidates}.

\paragraph{Interpreting the chronological trajectory.} The publication sequence shows methodological continuity with topical branching around a persistent tensor-network (TN) core. Early work is dominated by TN algorithms and quantum many-body applications. Later work carries the same representation and optimization language into TN machine learning, quantum-state and circuit construction, and quantum-AI-oriented directions \cite{ran2017criticality,zhou2021adqc,lu2023quvis,ran2023tnml}.

The three intervals contain 19 papers and 2 Hub births in 2010--2018, 18 papers and 10 births in 2019--2021, and 19 papers and 6 births in 2022--2026. Pooled canonical reuse is 0.211, 0.224, and 0.143 in these intervals, while mean old-node membership churn remains low throughout (0.00548, 0.00479, and 0.00378). Topic diversification helps explain the declining reuse trend, while the density outlier shows that compiler-level choices about semantic granularity also matter.

Network-based portraits of research have a long lineage. Author co-citation reveals intellectual structure, coauthorship graphs characterize collaboration, and maps of knowledge domains expose the organization of fields \cite{white1981author,newman2004coauthorship,borner2003visualizing}. Scientometric portraits have also combined productivity, collaboration, and citation indicators at the level of an individual researcher, while career-scale studies quantify the temporal evolution of scientific impact \cite{gonzalez2014portrait,sinatra2016impact,fortunato2018science}. ASKS brings this perspective to a source-compiled semantic graph. Branch survival, cross-paper support at birth, lineage branching, corpus coverage, and membership churn foreground the internal organization and evolution of a research program, complementing collaboration- and citation-based profiles.

\begin{figure*}[!tbp]
    \centering
    \safeincludegraphics[width=0.98\textwidth]{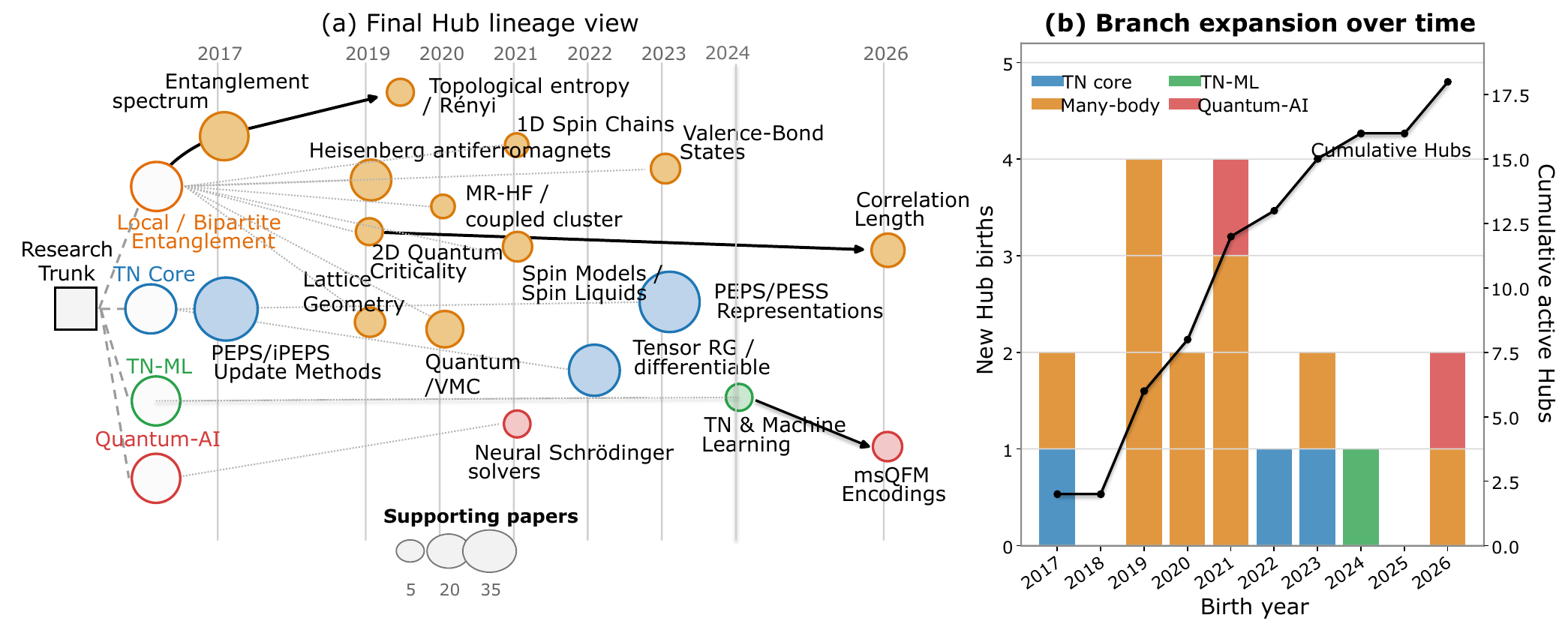}
    \caption{\textbf{A source-traceable author research portrait compiled from 56 publications.} (a) Topology-focused view of the final research tree. Vertical guides indicate Hub birth year, with small horizontal offsets used for readability. Colored anchors and light dotted connectors serve as display-level macro-domain aids. Solid black arrows show committed parent--child lineage, and node size encodes the final number of distinct supporting papers. The label msQFM denotes multi-spin quantum feature mapping. (b) Hub births by year and macro-domain together with the cumulative number of active Hubs. The strongest branching phase occurs in 2019--2021, while later births extend the TN core in 2023, TN-ML in 2024, and quantum-AI in 2026. The deterministic paper-to-Hub navigation mapping covers 50 of 56 papers and records six explicit abstentions. Hub birth, membership, lineage, and supporting-paper counts come from the frozen graph record. Macro-domain labels are post-hoc display interpretations, not generated Hub nodes, and do not enter the quantitative analysis.}
    \label{fig:portrait}
\end{figure*}

\paragraph{Author research portrait.} Because the corpus follows one researcher, the final Hub lineage yields a source-traceable \emph{author research portrait}. Figure~\ref{fig:portrait} shows committed Hub births and parent--child lineage from the frozen lifecycle record, with display-level macro-domains added for readability. The most persistent Hubs concern TN representations, PEPS/iPEPS, and tensor-renormalization methods. Later branches are associated with quantum many-body applications, TN-based machine learning, and quantum-AI-oriented work. Mapped quantum-computing papers are distributed across the TN core and adjacent branches, and six papers remain outside the final Hub mapping. The resulting tensor-network-centered research tree emphasizes methodological continuity more strongly than field labels alone would.

We quantify the tree shape with six descriptors. Let $N_{\mathrm{birth}}$, $N_{\mathrm{split}}$, $N_{\mathrm{merge}}$, and $N_{\mathrm{retire}}$ be the numbers of committed Hub lifecycle events in the chronological run. Let $N_{\mathrm{alive}}^{\mathrm{birth}}$ be the number of birth Hubs that remain active at $G_{56}$. Let $N_{\mathrm{multi}}^{\mathrm{birth}}$ be the number of multi-work birth events. Let $N_{\mathrm{lineage}}$ be the number of committed parent--child lineage links among births. Let $N_{\mathrm{mapped}}$ be the number of papers with a primary Hub assignment under the deterministic final-graph mapping. We report
\begin{equation}
\begin{aligned}
P_{\mathrm{survive}}&=N_{\mathrm{alive}}^{\mathrm{birth}}/N_{\mathrm{birth}},\\
B_{\mathrm{multi}}&=N_{\mathrm{multi}}^{\mathrm{birth}}/N_{\mathrm{birth}},\\
L_{\mathrm{branch}}&=N_{\mathrm{lineage}}/N_{\mathrm{birth}},
\end{aligned}
\end{equation}
\begin{equation}
\begin{aligned}
A_{\mathrm{life}}&=\frac{N_{\mathrm{birth}}}{N_{\mathrm{birth}}+N_{\mathrm{split}}+N_{\mathrm{merge}}+N_{\mathrm{retire}}},\\
Q_{\mathrm{map}}&=N_{\mathrm{mapped}}/56,\qquad
\bar C=\frac{1}{55}\sum_{t=2}^{56}C(t).
\end{aligned}
\end{equation}
These quantities characterize branch survival, cross-paper support, hierarchical refinement, lifecycle asymmetry, mapping coverage, and membership churn in the frozen trajectory.

\begin{table}[!tbp]
\centering
\footnotesize
\setlength{\tabcolsep}{3pt}
\caption{Graph descriptors of the chronological research portrait. Definitions are given in Eqs.~(6)--(7).}
\label{tab:portraitmetrics}
\begin{tabular}{@{}p{0.35\columnwidth}p{0.15\columnwidth}p{0.40\columnwidth}@{}}
\toprule
Descriptor & Value & Reading \\
\midrule
Survival $P_{\mathrm{survive}}$ & $1.000$ & All born branches persist to $G_{56}$. \\
Multi-work birth $B_{\mathrm{multi}}$ & $0.889$ & Most births are already cross-paper. \\
Lineage branching $L_{\mathrm{branch}}$ & $0.222$ & Some births refine earlier Hubs. \\
Lifecycle asymmetry $A_{\mathrm{life}}$ & $1.000$ & Observed growth is purely additive. \\
Mapping coverage $Q_{\mathrm{map}}$ & $0.893$ & 50 of 56 papers map into the final tree. \\
Mean old-node churn $\bar C$ & $0.00467$ & Membership is low-churn under the frozen configuration. \\
\bottomrule
\end{tabular}
\end{table}

Together, $P_{\mathrm{survive}}=1$ and $A_{\mathrm{life}}=1$ identify a persistent, birth-driven tree. The value $B_{\mathrm{multi}}=0.889$ shows that most branches begin with cross-paper support. The value $L_{\mathrm{branch}}=0.222$ captures selective hierarchical refinement. The combination of $Q_{\mathrm{map}}=0.893$ and $\bar C=0.00467$ indicates broad corpus coverage with low membership churn.

The mapping is designed for navigation. It records six abstentions at final Hub-overlap evaluation, and deterministic specificity and lineage rules resolve tied maximum overlap for 15 of the 50 mapped papers. These boundary cases identify where the compiled map abstains or applies an explicit navigation convention.

\subsection{What the Demonstration Shows}
Graph emergence in ASKS describes organization within the compiled knowledge system. A new Hub may reflect a genuine research theme, a methodological connection, terminology, or corpus bias. ASKS contributes an observable construction history: which sources entered, which local deltas were fused, how memberships changed, and which lineage constraints shaped later organization. Scientific interpretation begins with that trace and the source records it connects.

The author-corpus demonstration illustrates this relationship. Its Hub lineage is consistent with a recognizable research program with a stable tensor-network methodological trunk and later topical branching. Its evidential scope is one worked author corpus. Hub birth, membership, lineage, and supporting-paper counts are compiler outputs. The display-level macro-domains in Figure~\ref{fig:portrait} are post-hoc scientific interpretations that provide orientation. The demonstrated claim is structural: ASKS preserves enough continuity for a longitudinal research trajectory to become visible in graph state.

The resulting organization supports inspection, comparison, renaming, refinement, and investigation of Hubs through the underlying source record and compilation history. Scientific authority remains with those records and their assessment.

The present demonstration characterizes the frozen manifest trajectory. Permuted-order reconstruction and semantic-null controls are future validation tests for robustness and causal decomposition.

\section{Reliability and Bounded Authority}
Persistent knowledge becomes scientifically useful when graph growth can be distinguished from software drift. ASKS therefore treats reliability as part of the scientific architecture. Source records remain preserved through compilation. Wiki and semantic outputs have separate validation contracts. Graph changes are assembled as document-local deltas. Hard fusion violations trigger rollback at the document boundary. Reused relations accumulate source origins within compact topology. Research-state objects retain provisional status. Idempotence gives the same operation and inputs the same nodes, edges, origins, and memberships across repetitions.

Replay is the analogue of rerunning an experimental protocol. A replayable construction binds the source manifest, local model outputs, embedding identifier, configuration, operation versions, and ordered construction calls. The author-corpus demonstration applies this discipline through frozen local bundles, an isolated clean-state graph, hash-bound Hub decisions, step reports, and snapshot receipts. Supplementary Sections~\ref{sec:supp-repro}--\ref{sec:supp-engineering} provide the complete engineering record.

Agents remain useful within bounded authority. Mechanically checkable and rollback-safe operations can be automated. Changes that assign scientific meaning, merge identities, or approve a new Hub scope require an explicit gate. Concise summaries show the scientist which source changed the map, which Hub appeared, which members moved, and what evidence paths remain available. Detailed vector scores and engineering records remain available for audit in the supplement.

\section{Conclusion}
Persistent scientific knowledge construction can be treated as source-to-graph compilation. ASKS implements this process through source-local LLM encoding, validated document-local GraphDeltas, embedding-based cross-source organization, and deterministic graph rules for persistent state. The accepted state transitions retain their source links and construction history for replay and revision.

In the 56-paper chronological demonstration, compilation produces a low-churn Hub organization and a source-traceable author research portrait with explicit mapping coverage. The portrait has a tensor-network methodological core and later branches into several application domains. Higher-level organization is stable along the frozen trajectory. Canonical-node growth remains predominantly additive.

ASKS treats graph emergence as source-linked organization. Scientific novelty remains a separate review judgment. Its contribution is a construction process in which graph state can evolve while compiled views and navigation paths remain traceable to preserved scientific records.

\section*{Acknowledgments}
This work was supported in part by the National Natural Science Foundation of China (Grant No. 52202043), the Strategic Priority Research Program of the Chinese Academy of Sciences (Grant No.~XDB1270000) and the robotic AI-Scientist platform of the Chinese Academy of Sciences. Wen-Jun Li was supported by the Natural Science Foundation of Fujian Province under Grant No. 2026J0011096, the Startup Fund for Advanced Talents of Putian University under Grant 2024143, and the Putian Science and Technology Plan Project under Grant 2023GJGZ003.

\section*{Code and Data Availability}
The public ASKS source associated with this manuscript is available in the \href{https://github.com/ranshiju/Ran-ASKS/releases/tag/v0.2.0}{Ran-ASKS v0.2.0 release}. The same release contains frozen paper artifact 1.0.0: 56 sanitized paper Wiki pages, 18 Hub pages, the portable final graph, the reviewed corpus manifest, reported metrics, author-portrait tables, validation records, code provenance, and verification checksums. This release boundary distributes compiled and sanitized records. Primary PDFs, parsed Raw text, credentials, and private production knowledge-base state remain outside the public artifact. The software is released under the PolyForm Noncommercial License 1.0.0, and the frozen paper data and compiled Wiki/graph artifact are released under CC BY-NC 4.0. Subsequent ASKS development continues on the repository's main branch, while the results reported here remain bound to v0.2.0 and artifact 1.0.0.

\begingroup
\small
\balance

\endgroup

\clearpage
\onecolumn
\appendix
\setcounter{figure}{0}
\renewcommand{\thefigure}{S\arabic{figure}}
\begin{center}
{\LARGE\bfseries Supplementary Material}\par
\vspace{0.5em}
{\large LLMs Interpret, Embeddings Organize, Graphs Emerge:\par
Agent-Driven Compilation of Scientific Knowledge}
\end{center}
\vspace{1.5em}

\begin{center}
{\Large\bfseries Supplementary Contents}
\end{center}
\vspace{0.75em}

\newcommand{\suppcontentsline}[2]{%
  \noindent\hyperref[#1]{#2}\dotfill\hyperref[#1]{\pageref*{#1}}\par\vspace{0.45em}%
}
\suppcontentsline{sec:supp-repro}{A\quad Frozen Demonstration Boundary and Reproducibility Record}
\suppcontentsline{sec:supp-thresholds}{B\quad Frozen Organization Rules}
\suppcontentsline{sec:supp-state-machine}{C\quad Complete Ingest State Machine and Transaction Boundary}
\suppcontentsline{sec:supp-reingest}{D\quad Re-Ingest, Recompilation, and Implementation Boundaries}
\suppcontentsline{sec:supp-engineering}{E\quad Replay, Impact Analysis, and Construction Memory}
\suppcontentsline{sec:supp-manifest}{F\quad Frozen Publication Manifest}
\suppcontentsline{sec:supp-references}{Supplementary References}

\section{Frozen Demonstration Boundary and Reproducibility Record}
\label{sec:supp-repro}

The author-corpus demonstration was executed as a physically isolated construction run beginning from empty derived state. The frozen corpus contains 56 independently published works selected from 65 candidate records, together with three excluded candidates and six related-only records. A complete canonical PDF was manually verified for every included work. Formal journal issue year determines chronological grouping when an online-first date falls in the preceding calendar year. This convention places the boundary-singularity work with Run ID D046 in the 2024 group and the graph-learning work with Run ID D051 in the 2026 group. The frozen parse metadata records MinerU as the preferred structured-extraction engine for all 56 included PDFs \cite{wang2024mineru}. The resulting machine-readable derivatives were hash-bound before semantic compilation.

Construction was separated into two locked phases. In Phase L, each paper was freshly extracted and compiled into a Wiki view, semantic slots, validation reports, and a source-local bundle. The 56 bundles were structurally audited, semantically sampled, and frozen before graph-dependent work began. The run lock records DeepSeek AI's DeepSeek-V4-Flash-0731 for bibliographic review \cite{deepseek2026v4flash}, MiniMax-M3 with a DeepSeek fallback for local Wiki generation \cite{lai2026minimax}, and Zhipu AI Embedding-3 (\texttt{embedding-3}, recorded in the frozen run as GLM-Embedding-3) for semantic geometry \cite{zhipu2026embedding3}. A SHA-256 digest represents the API endpoint, while credentials remain in secure runtime configuration.

In Phase G, the frozen bundles were fused from an empty state $G_0$ in the frozen manifest order, grouped by formal publication year. The fusion lock binds the Phase-L run lock, both audit reports, all bundle hashes, the embedding identifier, the threshold configuration, graph schema, Hub-gate contract, relevant code hashes, runtime, and isolated write paths. Its SHA-256 digest is
\begin{quote}
\ttfamily\small
5ea7884df2255efddc92c9badd79efde8804ba58793bf55ba0e855e6ea5df2c0.
\end{quote}
The runtime used Python 3.12.13 on arm64 macOS with random seed 0. An isolated write allowlist targeted the experimental graph and Wiki paths. Every Hub-birth candidate received a hash-bound \texttt{commit\_birth} or \texttt{reject\_birth} decision through an explicit decision gate.

All 56 fusion steps and all 57 snapshots $G_0,\ldots,G_{56}$ completed. The 57 snapshot databases passed read-only integrity checks, 170 artifact receipts matched their recorded SHA-256 values, and all 56 step reports passed graph validation. These checks establish the identity and internal consistency of the reported construction trajectory within its frozen author-corpus scope.

\begin{figure}[htbp]
\centering
\safeincludegraphics[width=0.88\textwidth]{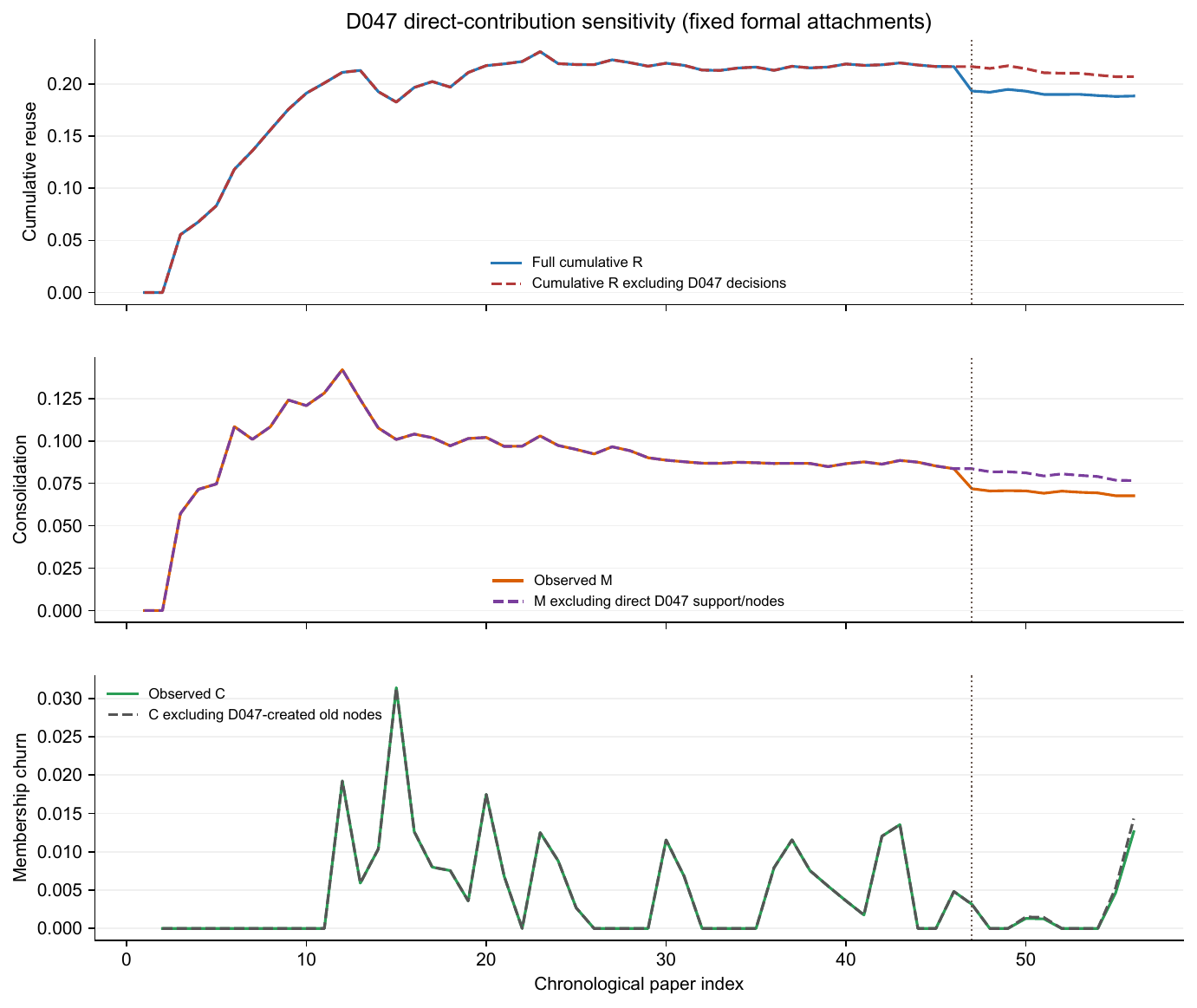}
\caption{\textbf{Direct-contribution sensitivity for the 2024 density-outlier paper (D047).} Solid curves show the observed frozen trajectory. Dashed curves remove D047's direct attachment, support, or created-old-node contribution under the fixed formal outcomes. The upper panel reports cumulative reuse $R_{\mathrm{cum}}(t)$. The comparison is deterministic and preserves all downstream fusion decisions.}
\label{fig:supp-sensitivity}
\end{figure}

The 2024 density-outlier paper, D047, is ``Eigenstate thermalization and its breakdown in quantum spin chains with inhomogeneous interactions'' (DOI: \href{https://doi.org/10.1103/PhysRevB.109.045139}{10.1103/PhysRevB.109.045139}). It contributes 105 create decisions and 2 reuse decisions. Figure~\ref{fig:supp-sensitivity} evaluates its direct contribution while retaining the fixed downstream fusion outcomes. Excluding that contribution changes $R_{\mathrm{cum}}(56)$ from 0.18854 to 0.20718, final $M$ from 0.06764 to 0.07663, and final $C$ from 0.01261 to 0.01432. The comparison quantifies leverage within the frozen trajectory and leaves the observed Hub lifecycle unchanged.

\section{Frozen Organization Rules}
\label{sec:supp-thresholds}

Table~\ref{tab:supp-thresholds} reports the principal numerical gates frozen before Phase G. They are implementation parameters of the demonstrated compiler. Identity, lineage, and high-consequence graph writes combine each score crossing with additional evidence.

\begin{table}[htbp]
\centering
\small
\begin{tabular}{p{0.18\textwidth}p{0.49\textwidth}p{0.24\textwidth}}
\toprule
\textbf{Process} & \textbf{Frozen signal and gate} & \textbf{Compilation effect} \\
\midrule
Node identity & lexical identity evidence, label similarity $\geq0.92$, semantic similarity $\geq0.90$, combined gate $\geq0.91$, winner margin $\geq0.04$, candidate floor $0.72$ & reuse, abstain, or create a local node \\
Proposition alignment & cosine $>0.98$ for reuse, relation band $0.90$--$0.98$ & merge, relate, or remain distinct \\
Paper routing & positioning text versus active Hub scope, routing floor $0.50$ and margin $0.04$ & paper-to-Hub navigation relation \\
Hub membership & enter $0.66$, retain $0.58$, at most three Hubs, child preference $+0.05$ & rebuild membership with hysteresis and specificity bias \\
Hub overload & active membership count $>20$ & redistribute toward children or trigger split analysis \\
Hub birth & pairwise similarity $\geq0.82$, at least four members, mean cohesion $\geq0.60$ & create a hash-bound candidate for explicit review \\
Hub split & at least six members, two groups of at least three, member-centroid and child-scope cosine $<0.85$, routing success $\geq0.80$ and margin $\geq0.03$ & create child candidates and migrate memberships after approval \\
Hub merge & scope and member-prototype similarity $\geq0.88$, distinct recent lineages & propose a lineage-preserving merge of distinct branches \\
\bottomrule
\end{tabular}
\caption{Frozen organization parameters used in the chronological demonstration.}
\label{tab:supp-thresholds}
\end{table}

\section{Complete Ingest State Machine and Transaction Boundary}
\label{sec:supp-state-machine}

A complete ingest is one ordered compiler pass:
\begin{enumerate}
    \item \textbf{Deduplicate and preprocess}: verify source identity and create an addressable machine-readable derivative when required.
    \item \textbf{Encode}: let the LLM produce the Wiki view and machine-facing semantic slots under separate output contracts.
    \item \textbf{Validate semantics}: normalize slots, enforce hard format constraints, and record or repair soft warnings.
    \item \textbf{Build the delta}: combine deterministic and semantic relations, construct the document-local GraphDelta, and plan attachment of boundary mentions.
    \item \textbf{Fuse}: apply the planned delta inside a database savepoint, recording new source origins on a single reusable edge.
    \item \textbf{Organize}: align propositions, refresh local Hub membership, apply specificity and overload rules, revisit abbreviations, and test birth, split, or merge candidates under lineage constraints.
    \item \textbf{Validate and finalize}: check graph invariants, record the ingest report, and update derived indexes and logs.
\end{enumerate}

Before fusion, every source must resolve to an addressable source package. Empty triples and self-loops are hard errors. Exact duplicates are removed. Boundary mentions pass through canonical identifiers, unique aliases, decomposed names, and finally an embedding-assisted identity gate. Ambiguity produces an explicit abstention, and persistent reuse requires sufficient identity evidence.

Fusion creates or reuses nodes according to the attachment plan and verifies that the page, source package, and source edge remain present. A hard-postcondition violation triggers complete rollback and restores the prior state. When an identical edge already exists, the canonical topology is reused and the new page is added to its origin record. Hub birth, split, and merge therefore operate downstream of source validation over a provenance-preserving graph.

\section{Re-Ingest, Recompilation, and Implementation Boundaries}
\label{sec:supp-reingest}

Re-ingest repeats source construction when a derivative, metadata record, or local semantic interpretation must be reconsidered. Recompilation rebuilds derived state from recorded sources and dependencies. Hub membership is rebuildable from current graph state. Split preserves the parent while creating children. Merge preserves the retired Hub through a redirect. These contracts allow active navigation to change while preserving both the source record and construction history.

Lexical evidence provides fallback identity resolution during embedding-service outages, and current Hub memberships remain stable until service resumes. The active compiler uses local attachment and Hub rules. An earlier production full-graph keyword similarity sweep belongs to a separate historical path, while the isolated author-corpus trajectory draws its evidence from the active compiler outputs.

Idempotence remains an engineering contract for re-ingest: repeating the same operation with the same inputs reproduces the same nodes, edges, origins, and memberships. A replay that changes the source bundle, model output, embedding identifier, schema, thresholds, or relevant code hashes receives a new lock and begins from clean derived state.

The production graph-governance layer also adapts Semantica's declarative-constraint, provenance, and temporal-validity patterns \cite{semantica2026repo} within the source-linked graph and its single-store boundary. These mechanisms strengthen validation and historical accountability while preserving the Raw--Wiki--graph evidence architecture used throughout ASKS.

\section{Replay, Impact Analysis, and Construction Memory}
\label{sec:supp-engineering}

For an agent-built knowledge substrate, replay is analogous to rerunning an experimental protocol. Append-only event streams and reversible, dependency-aware effects provide useful engineering precedents \cite{deepseek2026harness,shi2026cordis}. The formal demonstration uses frozen artifacts and isolated snapshots as a complete replay boundary. In production, selected sessions can serve as golden replays before changes to prompts, models, schemas, or graph code are released.

The broader system maintains an engineering dependency graph that links capabilities and contracts to the code, validation rules, and graph surfaces they affect. It supports impact analysis and targeted regression selection in a physically separated engineering layer whose role remains part of the overall ASKS architecture. The scientific graph organizes source-linked knowledge, the engineering graph organizes system dependencies, and session logs record execution diagnostics.

\paragraph{Typed agent orchestration.}
ASKS is treated here as a complete scientific knowledge system whose central persistent substrate is the source-linked knowledge graph together with its Raw and Wiki evidence surfaces. Work states, capabilities, and tools form operational layers built on this substrate. They extend how scientists and agents construct, query, interpret, validate, and present the compiled knowledge. The three mechanisms are distinguished by lifetime and effect. A \emph{work state} carries the active task, project context, and current phase across a sustained activity. A \emph{capability} contributes a temporary instruction and policy package when a particular action begins. A \emph{tool} executes a bounded function through explicit arguments, guards, and structured results. The engineering dependency graph provides shared discovery and impact context for all three, while their typed entry points preserve the distinction between persistent context, composable guidance, and execution. Table~\ref{tab:supp-agent-mechanisms} lists representative mechanisms implemented in the current system.

\begin{table}[htbp]
\centering
\footnotesize
\setlength{\tabcolsep}{4pt}
\begin{tabular}{@{}p{0.11\textwidth}p{0.23\textwidth}p{0.25\textwidth}p{0.32\textwidth}@{}}
\toprule
\textbf{Mechanism} & \textbf{Implemented examples} & \textbf{Registration and discovery} & \textbf{Invocation and operational role} \\
\midrule
Work state &
\texttt{research}: project-scoped research context with status and research memory &
Task route and the \texttt{research} engineering capability package &
\cliopt{task research} restores project context and carries evidence, decisions, experiments, and manuscript state across the sustained activity \\
\addlinespace
Work state &
Staged \texttt{query} loop and transactional \texttt{ingest} state machine &
Task route, stage cards, transaction checkpoints, and resumable receipts &
\cliopt{query-stage} selects \texttt{start}, \texttt{evidence}, \texttt{continue}, or \texttt{answer}. Ingest stages advance after validation and checkpoint completion \\
\addlinespace
Capability &
\texttt{write/academic}: shared writing conventions plus the physics-manuscript overlay &
\texttt{write} capability profile with a discipline-specific overlay &
\cliopt{capability write} \cliopt{capability-profile academic} loads claim-strength, scope, notation, and manuscript-QA conventions while preserving the active research state \\
\addlinespace
Tool &
\texttt{lookup}, \texttt{relations}, \texttt{read-raw}, \texttt{recall}, and \texttt{remember} &
DSH \texttt{ToolRegistry} and the unified \texttt{wg.py} command surface &
Structured arguments and results execute graph navigation, source-addressed reading, and project-memory operations under evidence guards \\
\addlinespace
\bottomrule
\end{tabular}
\caption{Representative implemented work states, composable capabilities, and executable tools in the ASKS scientific knowledge system. The examples cover a selected set of major implemented mechanisms built on the source-linked knowledge substrate.}
\label{tab:supp-agent-mechanisms}
\end{table}

\paragraph{Guarded and recoverable artifact workflows.}
Registered tools expose typed argument schemas and structured results, with pre- and post-execution guards enforcing evidence, path, privacy, and transaction constraints appropriate to each tool. Long-running workflows use content-addressed checkpoints whose run keys bind the source, code or prompt/schema version, model, execution parameters, and privacy permissions. A completed unit is reused only when these bindings match. Partial units are retried, and a changed binding creates a new run. Receipts record status, hashes, and compact validation results while credentials remain in runtime configuration. This gives ingestion, replay, manuscript inspection, and other long-running artifact workflows a common recoverable execution pattern.

Long-lived construction also requires decision memory: why a predicate was normalized, why a Hub scope was approved, why a parser changed, or why a validation rule was tightened. Source provenance explains why a knowledge object exists. Engineering decision records explain why the compiler behaves as it does. High-consequence writes remain bounded by explicit gates even when an agent proposes the candidate action.

\section{Frozen Publication Manifest}
\label{sec:supp-manifest}

Table~\ref{tab:supp-manifest} lists the 56-work corpus in the frozen chronological order defined by the publication manifest. Works sharing a formal publication year retain the manifest's frozen within-year order, and the D labels record the resulting fusion sequence. Dates follow the formal-publication convention described in Supplementary Section~\ref{sec:supp-repro}. The DOI column incorporates bibliographically verified corrections made after the experimental manifest was frozen. A dash marks an empty verified DOI field.

\begingroup
\scriptsize
\setlength{\tabcolsep}{3pt}
\begin{longtable}{@{}p{0.06\textwidth} p{0.055\textwidth} p{0.565\textwidth} p{0.27\textwidth}@{}}
\caption{Frozen 56-work corpus in the chronological order defined by the publication manifest.}\label{tab:supp-manifest}\\
\toprule
\textbf{Run ID} & \textbf{Year} & \textbf{Title} & \textbf{DOI} \\
\midrule
\endfirsthead
\multicolumn{4}{l}{\tablename\ \thetable\ continued from previous page}\\
\toprule
\textbf{Run ID} & \textbf{Year} & \textbf{Title} & \textbf{DOI} \\
\midrule
\endhead
\midrule
\multicolumn{4}{r}{Continued on next page}\\
\endfoot
\bottomrule
\endlastfoot
D001 & 2010 & Phase transitions and thermodynamics of the two-dimensional Ising model on a distorted kagome lattice & \doi{10.1103/PhysRevB.82.134434} \\
D002 & 2011 & Linearized Tensor Renormalization Group Algorithm for the Calculation of Thermodynamic Properties of Quantum Lattice Models & \doi{10.1103/PhysRevLett.106.127202} \\
D003 & 2012 & Phase diagrams, distinct conformal anomalies, and thermodynamics of spin-1 bond-alternating Heisenberg antiferromagnetic chain in magnetic fields & \doi{10.1103/PhysRevB.85.134425} \\
D004 & 2012 & Optimized decimation of tensor networks with super-orthogonalization for two-dimensional quantum lattice models & \doi{10.1103/PhysRevB.86.134429} \\
D005 & 2013 & Honeycomb Heisenberg spin ladder: Unusual ground state and thermodynamic properties & \doi{10.1209/0295-5075/104/57009} \\
D006 & 2013 & Theory of network contractor dynamics for exploring thermodynamic properties of two-dimensional quantum lattice models & \doi{10.1103/PhysRevB.88.064407} \\
D007 & 2013 & Kosterlitz--Thouless phase transition and re-entrance in an anisotropic three-state Potts model on the generalized kagome lattice & \doi{10.1103/PhysRevE.87.032151} \\
D008 & 2014 & Featureless quantum spin liquid, $1/3$-magnetization plateau state, and exotic thermodynamic properties of the spin-$1/2$ frustrated Heisenberg antiferromagnet on an infinite Husimi lattice & \doi{10.1103/PhysRevB.89.054426} \\
D009 & 2016 & Ab initio optimization principle for the ground states of translationally invariant strongly correlated quantum lattice models & \doi{10.1103/PhysRevE.93.053310} \\
D010 & 2016 & Phase diagram and exotic spin-spin correlations of anisotropic Ising model on the Sierpi\'nski gasket & \doi{10.1140/epjb/e2015-60745-5} \\
D011 & 2017 & Efficient perturbation theory to improve the density matrix renormalization group & \doi{10.1103/PhysRevB.95.064110} \\
D012 & 2017 & Fermionic algebraic quantum spin liquid in an octa-kagome frustrated antiferromagnet & \doi{10.1103/PhysRevB.95.075140} \\
D013 & 2017 & Criticality in two-dimensional quantum systems: Tensor network approach & \doi{10.1103/PhysRevB.95.155114} \\
D014 & 2017 & Few-body systems capture many-body physics: Tensor network approach & \doi{10.1103/PhysRevB.96.155120} \\
D015 & 2017 & Exploring Interacting Topological Insulators with Ultracold Atoms: The Synthetic Creutz--Hubbard Model & \doi{10.1103/PhysRevX.7.031057} \\
D016 & 2018 & Thermodynamics of spin-$1/2$ Kagom\'e Heisenberg antiferromagnet: algebraic paramagnetic liquid and finite-temperature phase diagram & \doi{10.1016/j.scib.2018.11.007} \\
D017 & 2018 & Emergent spin-1 trimerized valence bond crystal in the spin-$1/2$ Heisenberg model on the star lattice & \doi{10.1103/PhysRevB.97.075146} \\
D018 & 2018 & Controlling the phase diagram of finite spin-$1/2$ chains by tuning the boundary interactions & \doi{10.1103/PhysRevB.98.085111} \\
D019 & 2018 & Exotic entanglement scaling of Heisenberg antiferromagnet on honeycomb lattice & \doi{10.1140/epjb/e2018-90197-2} \\
D020 & 2019 & Machine learning by unitary tensor network of hierarchical tree structure & \doi{10.1088/1367-2630/ab31ef} \\
D021 & 2019 & Time matrix product state: theory and applications & \doi{10.7523/j.issn.2095-6134.2019.06.000} \\
D022 & 2019 & Noise-tolerant signature of $Z_N$ topological order in quantum many-body states & \doi{10.1103/PhysRevB.99.195101} \\
D023 & 2019 & Efficient quantum simulation for thermodynamics of infinite-size many-body systems in arbitrary dimensions & \doi{10.1103/PhysRevB.99.205132} \\
D024 & 2020 & Deep-neural-network solution of the electronic Schr\"odinger equation & \doi{10.1038/s41557-020-0544-y} \\
D025 & 2020 & Encoding of matrix product states into quantum circuits of one- and two-qubit gates & \doi{10.1103/PhysRevA.101.032310} \\
D026 & 2020 & Reentrance of the topological phase in a spin-1 frustrated Heisenberg chain & \doi{10.1103/PhysRevB.101.045133} \\
D027 & 2020 & Generative tensor network classification model for supervised machine learning & \doi{10.1103/PhysRevB.101.075135} \\
D028 & 2020 & Tangent-space gradient optimization of tensor network for machine learning & \doi{10.1103/PhysRevE.102.012152} \\
D029 & 2020 & Tensor network compressed sensing with unsupervised machine learning & \doi{10.1103/PhysRevResearch.2.033293} \\
D030 & 2021 & Deep Learning Quantum States for Hamiltonian Estimation & \doi{10.1088/0256-307X/38/11/110301} \\
D031 & 2021 & Automatically differentiable quantum circuit for many-qubit state preparation & \doi{10.1103/PhysRevA.104.042601} \\
D032 & 2021 & Preparation of many-body ground states by time evolution with variational microscopic magnetic fields and incomplete interactions & \doi{10.1103/PhysRevA.104.052413} \\
D033 & 2021 & Visualizing quantum phases and identifying quantum phase transitions by nonlinear dimensional reduction & \doi{10.1103/PhysRevB.103.075106} \\
D034 & 2021 & Phase identification in many-body systems by virtual configuration binarization & \doi{10.1103/PhysRevE.103.013313} \\
D035 & 2021 & Experimental realization of a quantum image classifier via tensor-network-based machine learning & \doi{10.1364/PRJ.434217} \\
D036 & 2021 & Predicting quantum potentials by deep neural network and Metropolis sampling & \doi{10.21468/SciPostPhysCore.4.3.022} \\
D037 & 2021 & Entanglement-Based Feature Extraction by Tensor Network Machine Learning & \doi{10.3389/fams.2021.716044} \\
D038 & 2022 & Unsupervised Recognition of Informative Features via Tensor Network Machine Learning and Quantum Entanglement Variations & \doi{10.1088/0256-307X/39/10/100701} \\
D039 & 2022 & Efficient simulation of quantum many-body thermodynamics by tailoring a zero-temperature tensor network & \doi{10.1103/PhysRevB.105.155155} \\
D040 & 2022 & Functional tensor network solving many-body Schr\"odinger equation & \doi{10.1103/PhysRevB.105.165116} \\
D041 & 2022 & Non-Parametric Semi-Supervised Learning in Many-Body Hilbert Space with Rescaled Logarithmic Fidelity & \doi{10.3390/math10060940} \\
D042 & 2023 & Nature of the $1/9$-magnetization plateau in the spin-$1/2$ kagome Heisenberg antiferromagnet & \doi{10.1103/PhysRevB.107.L220401} \\
D043 & 2023 & Tensor Network Efficiently Representing Schmidt Decomposition of Quantum Many-Body States & \doi{10.1103/PhysRevLett.131.020403} \\
D044 & 2023 & Quantum compiling with a variational instruction set for accurate and fast quantum computing & \doi{10.1103/PhysRevResearch.5.023096} \\
D045 & 2023 & Residual matrix product state for machine learning & \doi{10.21468/SciPostPhys.14.6.142} \\
D046 & 2024 & Boundary-induced singularity in strongly-correlated quantum systems at finite temperature & \doi{10.1088/2058-9565/ad038a} \\
D047 & 2024 & Eigenstate thermalization and its breakdown in quantum spin chains with inhomogeneous interactions & \doi{10.1103/PhysRevB.109.045139} \\
D048 & 2024 & Persistent Ballistic Entanglement Spreading with Optimal Control in Quantum Spin Chains & \doi{10.1103/PhysRevLett.133.070402} \\
D049 & 2025 & Entanglement scaling and criticality of infinite-size quantum many-body systems in continuous space addressed by a tensor network approach & \doi{10.1103/PhysRevB.111.245109} \\
D050 & 2025 & Compressing Neural Networks Using Tensor Networks with Exponentially Fewer Variational Parameters & \doi{10.34133/icomputing.0123} \\
D051 & 2026 & Unveiling the nature of graphs through quantum graphon learning & \doi{10.1038/s41534-025-01141-7} \\
D052 & 2026 & Machine learning of chaotic characteristics in classical nonlinear dynamics using variational quantum circuit & \doi{10.1088/1674-1056/adeb5b} \\
D053 & 2026 & Matrix-Product Entanglement Characterizing the Optimality of State-Preparation Quantum Circuits & \doi{10.1103/6y5p-mp7q} \\
D054 & 2026 & Statistics-encoded tensor network approach in disordered quantum many-body spin chains & \doi{10.1103/vb4b-m5h6} \\
D055 & 2026 & Exotic transition and reentrance on the $1/9$ magnetization plateau in a spin-$1/2$ anisotropic kagome antiferromagnet & \doi{10.1103/vwx9-3zwj} \\
D056 & 2026 & Orthogonality in Quantum-Probabilistic Machine Learning: An Investigation on Multiqubit Encoding & \doi{10.34133/icomputing.0232} \\
\end{longtable}
\endgroup

\renewcommand{\refname}{Supplementary References}

\end{document}